# Wavelet Based Normal and Abnormal Heart Sound Identification using Spectrogram Analysis

Nilanjan Dey
Assistant Professor ,Department of IT
JIS College of Engineering
Kalyani, West Bengal, India.

Achintya Das
Professor and Head, ECE Department
Kalyani Govt. Engineering College
Kalyani, West Bengal, India.

Sheli Sinha Chaudhuri
Associate Professor, ETCE Department
Jadavpur University
Kolkata, West Bengal ,India

*Abstract –* **The present work proposes a computer-aided normal and abnormal heart sound identification based on Discrete Wavelet Transform (DWT), it being useful for tele-diagnosis of heart diseases. Due to the presence of Cumulative Frequency components in the spectrogram, DWT is applied on the spectrogram up to** *n* **level to extract the features from the individual approximation components. One dimensional feature vector is obtained by evaluating the Row Mean of the approximation components of these spectrograms. For this present approach, the set of spectrograms has been considered as the database, rather than raw sound samples. Minimum Euclidean distance is computed between feature vector of the test sample and the feature vectors of the stored samples to identify the heart sound. By applying this algorithm, almost 82% of accuracy was achieved.**

I. INTRODUCTION

Heart Auscultation is used for interpreting acoustic waves produced by the mechanical action of the heart [3]. It is a screening method that is used as a primary tool in the diagnosis of cardiac diseases. This interpretation can provide valuable information regarding the function of heart valves, and is capable of detecting various heart disorders, especially problems related to the valves. Auscultation [1, 2] comprises of two phases: heart sound acquisition and heart sound analysis. Heart sound acquisition involves placing an electronic stethoscope at the apt location on a patient's chest coupled with the right amount of pressure to capture heart sound. Electronic Stethoscope is designed in such a way that it can transfer these signals to the associated computer unit using wireless media. Heart sound analysis helps in determining whether the captured sound corresponds to a healthy or a diseased heart.

In healthy adults, heart sounds consists of mainly two events: the first heart sound ($S_1$) and the second heart sound ($S_2$). Together they are referred to as *Fundamental Heart Sound* (FHS) [4]. A cardiac cycle or a single heartbeat is the interval between the commencements of the first heart sounds till the commencement of the immediate first heart sound. *Systole* is defined as the interval between the ends of first heart sound to the commencement of the same cycle's second heart sound. The interval between the ends of second heart sound to the commencement of the next cycle's first heart sound is called *diastole*. Generally, systole is shorter than diastole. Apart from FHS, there can be a triple rhythm in diastole called a *gallop*, resulting in the presence of third heart sound ($S_3$), fourth heart sound ($S_4$) or both.

Cardiac cycles of abnormal heart sounds contains some special components. These components can be classified into two types: extra heart sounds and murmur sounds. An example of extra heart sound is the third heart sound. The second type of abnormal heart sound is a murmur, which is caused by turbulent blood flow through a blocked (stenosis) valve, or backward flow through a leaking (regurgitation) valve. These sounds can be heard in both systole and diastole. The occurrence of murmur is a good indicator of valve related disorders.

The present work proposes a method for normal and abnormal heart sound identification based on spectrogram analysis using Discrete Wavelet Transform and Euclidean Distance Measurement between the feature vectors of test and training Heart sound samples.

## II. METHODOLOGY

Spectrogram [5, 6] is used since long time for 1D signal recognition. Spectrogram is a time-varying spectral representation that plots the variation of spectral density with respect to time. Spectrogram is a two dimensional graph, where horizontal axis represents time and vertical axis represents frequency. A third dimension indicating amplitude of a particular frequency is represented by the intensity or color of each point in the image. For analog 1D signal processing, approximation as a filter bank that results from a series of band pass filters and calculation from the time signal using the Short-Time Fourier Transform (STFT) are adopted for spectrogram analysis. But due to more precise and optimized calculation this approach is done based on digital 1D signal processing using DWT.

### A. Discrete Wavelet Transform (DWT)

The wavelet transform describes a multi-resolution decomposition process in terms of expansion of an image onto a set of wavelet basis functions. Discrete Wavelet Transformation has its own excellent space frequency localization property. Application of DWT [7] in 2D images corresponds to 2D filter image processing in each dimension. The input image is divided into 4 non-overlapping multi-resolution sub-bands by the filters, namely LL1 (Approximation coefficients), LH1 (vertical details), HL1 (horizontal details) and HH1 (diagonal details). The sub-band (LL1) is processed further to obtain the next coarser scale of wavelet coefficients, until some final scale "N" is reached. When "N" is reached, 3N+1 sub-band are obtained consisting of the multi-resolution sub-bands. Which are LLX and LHX, HLX and HHX where "X" ranges from 1 until "N." Generally, most of the image energy is stored in the LLX sub-bands.

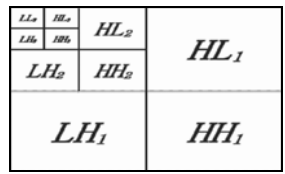

Figure 1 Three phase decomposition using DWT

The Haar wavelet is the simplest possible wavelet. Haar wavelet is not continuous, and therefore not differentiable. This property can, however, be an advantage for the analysis of signals with sudden transitions.

## III. PROPOSED METHOD

Step 1. Test heart sound sample is converted into spectrograms.
Step 2. The Discrete Wavelet Transform is applied up to level 4 on the spectrogram to obtain the feature vector by considering the row means of the absolute values of approximate components.
Step 3. Training sound samples are resized based on test sample.
Step 4. The Discrete Wavelet Transform is applied up to level 4 on the spectrogram to obtain the feature vector by considering the row means of the absolute values of approximate components of the trained samples.
Step 5. Euclidean distance is computed between the test sample feature vector and the entire training sample feature vectors.
Step 6. The minimum distance between test and training sample feature vector is measured.

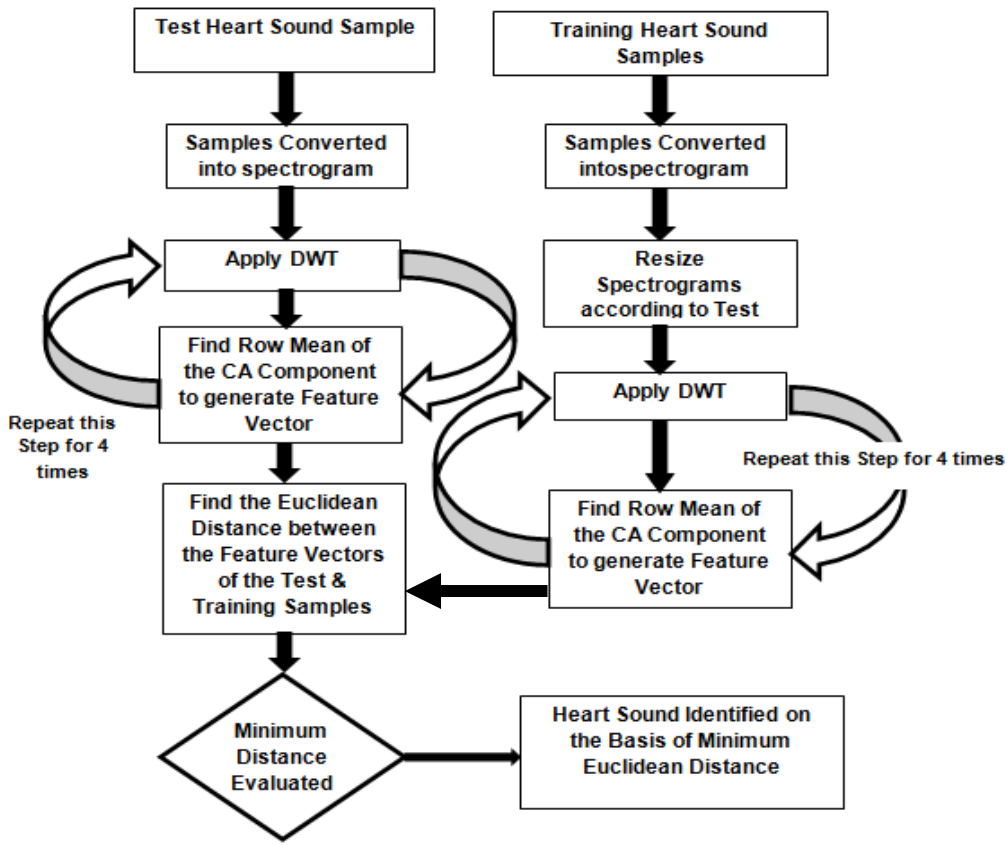

Figure 2 Heart Sound Identification Process

IV. EXPLANATION OF THE PROPOSED METHOD

In this proposed approach, the test heart sound and the training heart sound samples both are converted into spectrograms. These spectrograms of the training samples are then resized according to the test sample. The Discrete Wavelet Transform is then applied on both test and training sample up to the fourth level. For both test and training spectrograms, row means were calculated by considering the absolute values of individual rows of each approximate (Ca) component for different level of sub-band decomposition. After getting the row vector for each iteration, a one dimensional feature vector is formed. Euclidean distance is computed between test and each of the training samples.

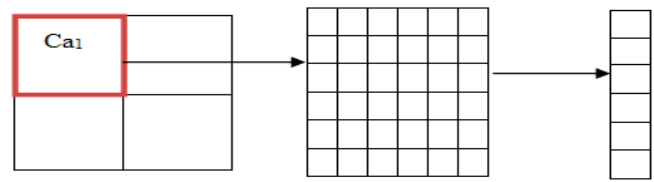

Figure 3 Feature vector (Row-Mean of abs (Ca (i, j))) generation after 1$^{st}$ level DWT decomposition

Euclidean distance between the points $X_{test}$ (X1, X2, etc.) and point $Y_{train}$ (Y1, Y2, etc.) is calculated by using the formula shown in equation

$$D = \sqrt{\sum_{i=1}^{n}(X_i - Y_i)^2} \quad \ldots(1)$$

Smallest Euclidean distance between test and training feature vectors signifies the identification of the type of heart sound.

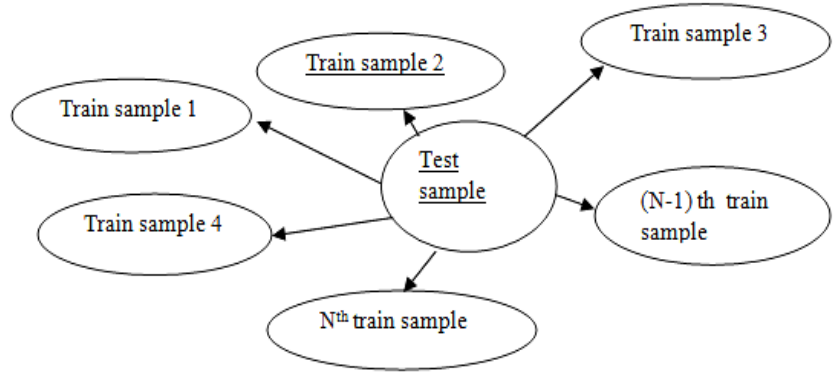

Figure 4 Heart Sound Matching Process

V. RESULT AND DISCUSSION

MATLAB 7.0.1 Software is extensively used for the study of the heart sound identification process. Concerned images obtained in the result are shown in Figure 5 through 13. The Dataset which is used for our study, containing 176 files in WAV format of Normal, Murmur and Extra Heart sound type [8].

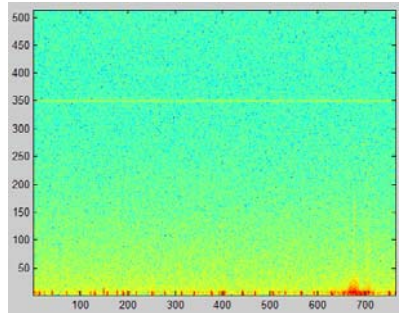

Figure 5 Spectrogram of the Normal Heart Sound

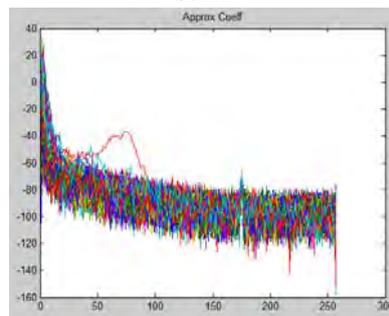

Figure 6 $Ca_1$ component after applying DWT on the spectrogram

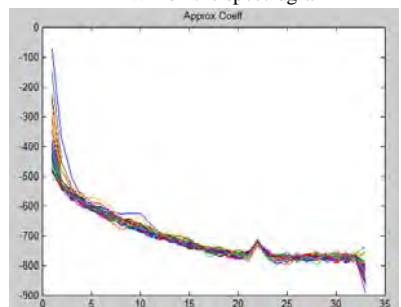

Figure 7 $Ca_4$ component after applying DWT on $Ca_3$.

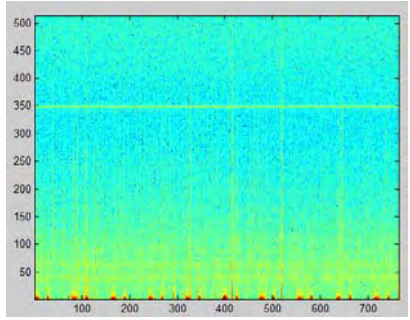

Figure 8 Spectrogram of a Murmur Sound

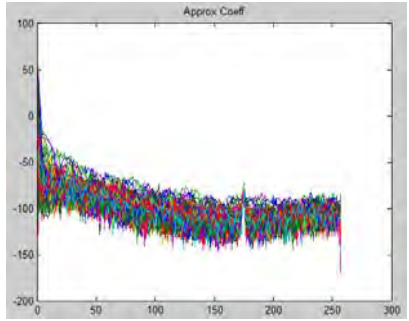

Figure 9 $Ca_1$ component after applying DWT on the spectrogram

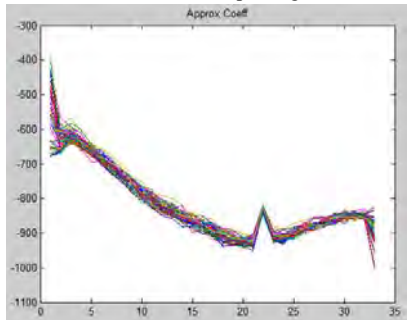

Figure 10 $Ca_4$ components after applying DWT on $Ca_3$

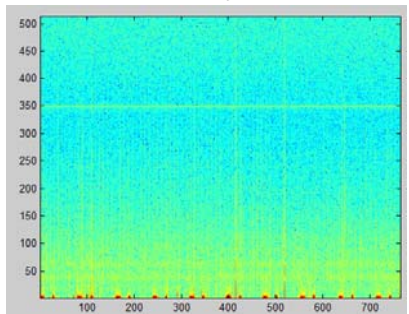

Figure 11 Spectrogram of an Extra Heart Sound

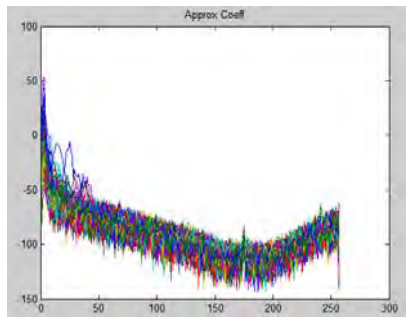

Figure 12 $Ca_1$ component after applying DWT on the spectrogram

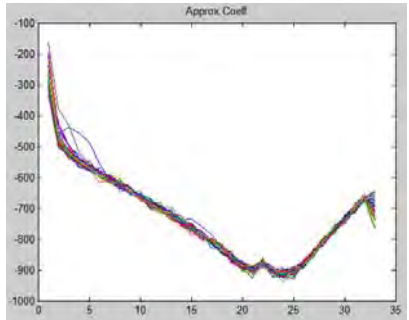

Figure 13 $Ca_4$ components after applying DWT on $Ca_3$

TABLE 1

Test Case 1

| Murmur Sound as a Test Sample Vs | Euclidean Distance | Match/Mismatch |
|---|---|---|
| Extra Heart Sound as Training Sample 1 | 929.3838 | Mismatched |
| Normal Heart sound as Training Sample 2 | 841.1171 | Mismatched |
| Murmur Sound as Training Sample 3 | 285.2517 | **Matched** |

TABLE 2

Test Case 2

| Extra Sound as Test Sample Vs | Euclidean Distance | Match/Mismatch |
|---|---|---|
| Extra Heart Sound as Training Sample 1 | 680.2363 | **Matched** |
| Normal Heart sound as Training Sample 2 | 837.9943 | Mismatched |
| Murmur Sound as Training Sample 3 | 929.3838 | Mismatched |

TABLE 3

Test Case 3

| Normal Sound as Test Sample Vs | Euclidean Distance | Match/Mismatch |
|---|---|---|
| Extra Heart Sound as Training Sample 1 | 1.3567e+003 | Mismatched |
| Normal Heart sound as Training Sample 2 | 1.1113e+003 | **Matched** |
| Murmur Sound as Training Sample 3 | 1.2807e+003 | Mismatched |

## VI. CONCLUSION

This present work dealt with 176 nos. of normal and abnormal heart sound samples. At the time of recognition process, each test sample took almost a minute to be identified properly. Though it was a heart sound recognition process, still the heart sound was converted into image form for proper identification of cumulative frequency components, and special features that are used to distinguish the proper identified heart sound from other training samples present in the database. Several iterations of row mean feature vector generation technique are very effective for complexity reduction. By applying this algorithm, almost 82% of accuracy was achieved. By applying some different feature extraction techniques as column mean, or both row and column mean, moving average technique etc. and further developments, this limitation is expected to be resolved.